\documentclass[10pt,twocolumn,letterpaper]{article}

\usepackage{cvpr}
\usepackage{times}
\usepackage{epsfig}
\usepackage{graphicx}
\usepackage{amsmath}
\usepackage{amssymb}
\usepackage{diagbox}
\newcommand\numberthis{\addtocounter{equation}{1}\tag{\theequation}}

\usepackage{soul}

\usepackage[pagebackref=true,breaklinks=true,letterpaper=true,colorlinks,bookmarks=false]{hyperref}

\cvprfinalcopy 

\def\cvprPaperID{3296} 

\begin{document}

\title{Contextual Visual Similarity}

\author{Xiaofang Wang, Kris M. Kitani and Martial Hebert\\
The Robotics Institute\\
Carnegie Mellon University\\
{\tt\small xiaofan2@andrew.cmu.edu \{kkitani,hebert\}@cs.cmu.edu}
}

\maketitle

\begin{abstract}
Measuring visual similarity is critical for image understanding. But what makes two images similar? Most existing work on visual similarity assumes that images are similar because they contain the same object instance or category. However, the reason why images are similar is much more complex. For example, from the perspective of category, a black dog image is similar to a white dog image. However, in terms of color, a black dog image is more similar to a black horse image than the white dog image. This example serves to illustrate that visual similarity is ambiguous but can be made precise when given an explicit contextual perspective. Based on this observation, we propose the concept of contextual visual similarity. To be concrete, we examine the concept of contextual visual similarity in the application domain of image search. Instead of providing only a single image for image similarity search (\eg, Google image search), we require three images. Given a query image, a second positive image and a third negative image, dissimilar to the first two images, we define a contextualized similarity search criteria. In particular, we learn feature weights over all the feature dimensions of each image such that the distance between the query image and the positive image is small and their distances to the negative image are large after reweighting their features. The learned feature weights encode the contextualized visual similarity specified by the user and can be used for attribute specific image search. We also show the usefulness of our contextualized similarity weighting scheme for different tasks, such as answering visual analogy questions and unsupervised attribute discovery.
\end{abstract}

\section{Introduction}

Measuring the visual similarity between images is an important aspect of artificial intelligence. But what makes two images similar? Most existing work on visual similarity has an underlying assumption about the way images are similar. Traditionally, two images are considered as visually similar if the two images contain the same object instance or objects of the same category, \eg, two pictures of a dog. However, the reason why images are similar is much more complex than just containing the same object instance or category. Alternatively, one could imagine defining a set of attributes over which images could be similar (\eg, color, texture, scene type, \etc.), but this approach is problematic as there are too many ways in which two images can be similar. Furthermore, it is not clear if one attribute should be more important that another attribute when defining similarity. To resolve this dilemma of multi-faceted similarity, we propose a new paradigm for specifying image similarity, where we define the similarity of two images given a third `context' image.

\begin{figure}
\centering
\includegraphics[width=0.9\columnwidth]{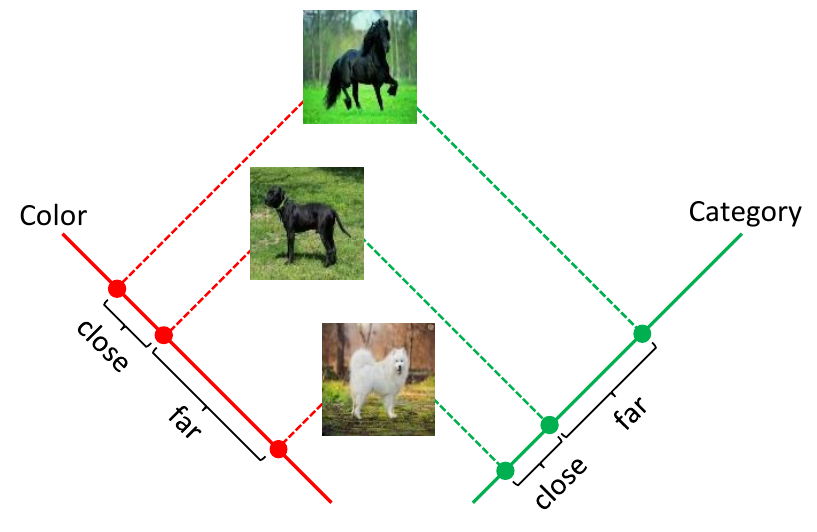}
\caption{An example that illustrates visual similarity between images is relative. In terms of category, the black dog is more similar to the white dog. However, in terms of color, the black dog is more similar to the black horse.}
\label{introexample}
\end{figure}

What makes visual similarity a complex concept is that there are many perspectives that can be used to determine the similarity between two images. For example, consider a black dog, a white dog and a black horse as shown in Figure~\ref{introexample}. In terms of object category, the black dog is more similar to the white dog than the black horse. But in terms of color, the black dog is more similar to the black horse instead of the white dog. The visual similarity between two images is relative, as it changes as we change our mental perspective, which we call this \textit{context}. This toy example suggests that it is important to make this context explicit when computing image similarities.

One possible way to address the above issue is to somehow model the user's mental perspective (context), \ie, infer the reason why two images should be similar. Inferring the context of visual similarity involves a deep understanding of images and is difficult even for humans. What makes the problem so challenging is that there is a set of equally valid contexts under which two images can be similar. For example, there are different, equally valid answers to the question that why the black dog image and the black horse image are similar, including that they are both black, furry, four-legged animals. Although the task of inferring the context of similarity is an interesting research question, we focus on the more constrained situation where we ask a human to provide the context in the form of an additional image.

In this work, we consider the special case in which the user provides a triplet of images to define the context of visual similarity. Specifically, the user supplies: (1) a query image; (2) a positive image similar to the query image; and (3) a negative image dissimilar to the first two images. We will show that the addition of this context can greatly help narrow down valid reasons for similarity. For example, if we are given (1) black dog (query); (2) white dog (positive) and (3) black horse (negative), we can infer that the similarity between the query and positive image is due to the object category and not the color. Given this set of image triplets as input, we show later how this paradigm can be used for a variety of image understanding tasks.

Concretely, given a query triplet (query, positive, negative image) we learn a reweighting of the image features such that the user-defined visual similarity relationship holds. We first represent images as features and then learn weights over each dimension to force that the query image and the positive image are close to each other but also far from the negative image. The learned feature weight vector encodes the contextual visual similarity -- the user's perception of similarity given context. 

An immediate advantage of such an approach is that now we can allow users to search for images similar according to a specific contextual visual similarity. Our approach can support the task of attribute specific image search without having to explicitly train attribute classifiers. Our approach is also useful for answering visual analogy questions which takes the form of \textit{image A is to image B as image C is to what}~\cite{sadeghi2015visalogy}. Our approach can also be used to perform unsupervised attribute discovery.

\textbf{Contributions:} Our main contribution is the idea of modeling contextual visual similarity. To this end, we propose to learn a feature weight vector to adapt the original feature to make a given contextual visual similarity relationship hold. We demonstrate the effectiveness of our approach on three tasks, including (1) attribute specific image search, (2) answering visual analogy questions, and (3) unsupervised attribute discovery.

\section{Related Work}

\noindent
\textbf{Relevance Feedback in Image Search:} Relevance feedback (RF) has been extensively used to improve interactive image search~\cite{ ferecatu2007interactive, rui1998relevance, tieu2004boosting, zhou2003relevance}. Users are asked to provide relevance scores for a set of initially retrieved images~\cite{rui1998relevance} or select images closest to the target image among displayed images~\cite{ferecatu2007interactive}. The user's feedback is then used to prune the initial search results. The application of our approach to attribute specific image search can be viewed one kind of relevance back, where the user proactively provides positive and negative images to guide the search. However, RF approaches usually involve multiple iterations of user feedback and result pruning, while we only require to the user to provide guidance at the time of initiating the search. Moreover, search examples provided in~\cite{ferecatu2007interactive} usually involve searching for images within the same category. In contrast, we are searching for images containing a user defined attribute.

\noindent
\textbf{Attributes for Image Search:} Recently, attributes have been employed to aid image retrieval~\cite{douze2011combining, kovashka2013attribute2, kovashka2013attribute, kovashka2012whittlesearch, kumar2008facetracer, rastegari2013multi, siddiquie2011image}. Visual attributes are used as features to represent images in~\cite{douze2011combining}. Attributes or relative attributes are also used to describe users' target visual content~\cite{kumar2008facetracer, rastegari2013multi, siddiquie2011image} or provide feedback to the search results~\cite{kovashka2013attribute2, kovashka2013attribute, kovashka2012whittlesearch}. While all of these approaches can search for images with specific attributes, they need users to explicitly name the target attribute and require attribute classifiers pre-trained on supervised attribute labels. Our approach only requires users to provide image examples and our approach does not require any explicit attribute labels.

\noindent
\textbf{Visual Analogy:} The problem of solving analogy questions has been well explored in NLP. For example, Latent Relational Analysis (LRA)  introduced in~\cite{turney2006similarity} can be used to solve SAT word analogy questions. In the visual domain, \cite{hwang2013analogy} proposes an analogy-preserving semantic embedding, which is proved useful for object categorization. \cite{hertzmann2001image} and \cite{reed2015deep} can synthesize an `analogous' image $D$ that relates to image $C$ in the same way as image $B$ related to image $A$. \cite{reed2015deep} demonstrates the effectiveness of their model on synthetic data, where images are controlled and visual analogies mainly involve relatively low-level visual properties such as rotation. Instead of synthesizing new images, VISALOGY~\cite{sadeghi2015visalogy} solves visual analogies by discovering the mapping from image A to image B and searching for an image D such that A to B holds for C to D. VISALOGY conducts experiments on natural image datasets, where visual analogies involve different high-level semantic properties. Our work is different in that we do not explicitly model the mapping between image A and B. Instead, we use cues derived from contextual visual similarity relationships to solve visual analogies.


\noindent
\textbf{Attribute Discovery:} Attribute discovery has been studied in recent work~\cite{berg2010automatic, rastegari2012attribute,shankar2015deep,vittayakorn2016automatic}. \cite{berg2010automatic} demonstrates that it is possible to identify attribute vocabularies and learn to recognize attributes automatically by mining text and image data collected from the Internet. \cite{vittayakorn2016automatic} proposes to automatically discover visual attributes from a noisy collection of image-text data by exploiting the relationship between attributes and neural activations in the deep network. \cite{shankar2015deep} proposes a novel training procedure with CNNs to discover multiple visual attributes in images in a weakly supervised scenario. All these works leverage textual description or partial attribute labels of images, whereas our approach does not require such side information. \cite{rastegari2012attribute} proposes a novel way of representing images as binary codes that balances discrimination and learnability of the codes. They show that their codes can be thought of as attributes. In contrast, we observe that one meaningful contextual visual similarity relationship entails semantic attributes and we propose to discover attributes by clustering contextual visual similarity relationships.

\noindent
\textbf{Representation Learning:} Our work is also related to previous work on representation learning~\cite{wang2014learning, wang2015unsupervised, schroff2015facenet, hoffer2015deep, veit2016disentangling}. These works perform representation learning with triplets of images. However, our work does not learn a new image representation. Instead, we learn feature weights for the existing image representation.  After modifying the representation by different weights, different visual properties (attributes) can be represented. The idea of reweighting image representation with different weights to represent different visual properties is also studied in~\cite{veit2016disentangling}. However, in their model, the weights are either pre-set or learned from a large number of triplets with labels. Their model can only handle visual properties with labels, which limits their model's applicability in our scenario. Our approach does not use any labels and has no restriction on visual properties.

\section{Approach}

We first present our approach for modeling contextual visual similarity. Then we explain how we apply our approach to the task of attribute specific image search, answering visual analogy questions and unsupervised attribute discovery.

\subsection{Modeling Contextual Visual Similarity}
\label{modelreason}

We are given a set of training images $\mathcal{I} = \{I_i\}$  represented by feature vectors $\{x_i = f(I_i)\} \in \mathbb{R}^d$. We denote a triplet of image indices by $(q, p, n)$, indicating that the query image of index $q$ and the positive image of index $p$ are similar to each other but both dissimilar to the negative image of index $n$.

Our goal is to learn a weighting on the image features such that $I_{q}$ and $I_{p}$ are similar to each other but dissimilar to $I_{n}$. In the original feature space, the Euclidean distance between $x_q$ and $x_p$ is not necessarily small and their Euclidean distances to $x_n$ are not necessarily large, \ie, the visual similarity relationship between $I_q$, $I_p$ and $I_n$ does not necessarily hold in the original feature space. We propose to learn feature weights $w \in \mathbb{R}^d$ over all the feature dimensions to ensure that $I_{q}$ and $I_{p}$ are close to each other but also far from $I_{n}$ after reweighting the features. Concretely, we define $W = \text{diag}(w_1, w_2, \ldots, w_d)$ and want the Euclidean distance between $W x_{q} $ and $W x_{p} $ to be small and their Euclidean distances to $W  x_{n}$ to be large. After reweighting features with $w$, $I_q$ and $I_p$ will be close to each other and both far from $I_n$, \ie, the visual similarity relationship between $I_{q}$, $I_{p}$ and $I_{n}$ holds due to the learned feature weight vector $w$. In another word, $w$ encodes the contextual visual similarity relationship between $I_{q}$, $I_{p}$ and $I_{n}$.

\subsubsection{Learning the Feature Weights}

In this section, we describe how we learn the feature weight vector $w$. We start by defining a triplet ranking loss function and show how we adapt it according to our scenario. The triplet ranking loss function is defined as follows
\begin{align*}
L_t = &\max\{0, ||Wx_q - Wx_p||^2_2 - ||Wx_q - Wx_n||^2_2 + \alpha\},\numberthis \label{eq1}
\end{align*}
where $\alpha$ denotes the margin, $||\cdot||_2$ denotes the $L_2$ norm of vectors and $W = \text{diag}(w_1, w_2, \ldots, w_d)$. But directly using the above loss function may lead to overfitting. The query image may be pushed too close to the positive image or too far from the negative image after reweighting features. Inspired by the contrastive loss with two margins proposed in~\cite{sadeghi2015visalogy}, we propose the following loss function with two margins:
\begin{align*}
L_t = &\max\{0, ||Wx_q - Wx_p||^2_2 - \alpha_p\}+\\
&\max\{\alpha_n - ||Wx_q - Wx_n||^2_2, 0 \},\numberthis \label{eq2}
\end{align*}
where $\alpha_p$ and $\alpha_n$ denote the positive margin and negative margin respectively. Using this loss function, the query image and the positive image will be pushed closer to each other only when the distance between them is larger than $\alpha_p$. Likewise, the query image and the negative image will be pushed further to each other only when the distance between them is smaller than $\alpha_n$. Note that, mathematically, the distance mentioned here is the square Euclidean distance between feature vectors. The two margins help regularize the feature space after reweighting and avoids overfitting.

The above loss function only ensures that the query image is far from the negative image after feature reweighting. We also want the positive image to be far from the negative image, so we modify the loss function in the following way:
\begin{align*}
L_t = &\max\{0, ||Wx_q - Wx_p||^2_2 - \alpha_p\}+ \\
&\max\{\alpha_n - ||Wx_q - Wx_n||^2_2, 0 \}+\\
&\max\{\alpha_n - ||Wx_p - Wx_n||^2_2, 0 \}.\numberthis \label{eq3}
\end{align*}
In the loss function in Eq.~\eqref{eq3}, we add constraints on the distance between positive and negative images.

We can also assume that the given feature vectors $\{x_i\}$ are unit vectors and we also hope that after reweighting, they are still unit vectors. To do this we define the following regularization term:
\begin{align*}
L_r = \sum_{i=p,q,n} (||Wx_i||_2^2 - 1)^2, \numberthis \label{eq4}
\end{align*}
and our final loss function becomes
\begin{equation}
L = L_t + \lambda L_r,
\end{equation}
where $\lambda$ is a hyper-parameter to balance the triplet loss and the regularization term.

We learn the feature weight vector $w$ by minimizing $L$. The loss function $L$ is differentiable with respect to $w$, thus gradient descent can be applied to minimize the loss function. We only show the loss function defined over a single triplet of images for brevity. This can easily be extended to a set of triplets.

\subsection{Attribute Specific Image Search}


Most existing search-by-example image search applications take one query image as input and retrieve images containing the same object instance or category with the query as they assume the user is considering instance-level or category-level image similarity. These applications do not allow a user to specify contextual information to better define similarity.

We show how our proposed approach can be used for the task of attribute specific image search, which aims to enable users to search for images containing a target attribute. In addition to providing one query image containing the target attribute, the user is also asked to provide a few examples of positive images, which contain the target attribute but in a different object category with the query. The user is also asked to provide negative images, which do not contain the target attribute but in the same object category with the query. Note that the user only needs provide images and does not need to name the target attribute explicitly.


Once we have the query, positive and negative images provided by the users, we can infer the weights of the contextual visual similarity measure. The contextual visual similarity measure can be used to guide the image search. In this way, we are able to represent the user's mental model better and return better search results.

More specifically, we form all possible (query, positive, negative) triplets with images provided by the user. Then, the feature weight vector $w$ is learned based on these triplets with our approach. Once $w$ is learned, we reweight the feature of the query and all the database images with $w$. Finally all the database images are ranked according to their distance to the query after feature reweighting.

\subsection{Answering Visual Analogy Questions}

The second task that modeling contextual visual similarity can help us with is answering visual analogy questions. These questions take the form of \textit{image A is to image B as image C is to what?}~\cite{sadeghi2015visalogy}. A general form of valid analogy quadruples can be written as follows
\begin{equation}
[I_1^{c_1,p_1}:I_2^{c_1,p_2}::I_3^{c_2,p_1}:I_4^{c_2,p_2}],
\label{quesform}
\end{equation}
where $I_1$ and $I_2$ belong to the same category $c_1$ but have different properties $p_1$ and $p_2$ respectively. Likewise,  $I_3$ and $I_4$ belong to the same category $c_2$, where $c_1 \neq c_2$, but have different properties $p_1$ and $p_2$. Example properties include color, action and object orientations~\cite{sadeghi2015visalogy}.

VISALOGY~\cite{sadeghi2015visalogy} poses answering a visual analogy question $I_1:I_2::I_3:?$ as the problem of discovering the mapping from image $I_1$ to image $I_2$ and searching for an image $I_4$ such that $I_1$ to $I_2$ holds for $I_3$ to $I_4$. Instead of discovering the mapping from $I_1$ to $I_2$, we use cues derived from contextual visual similarity relationship to solves visual analogy questions.

We observe two contextual visual similarity relationships from visual analogy questions. First, in terms of category $c_1$, $I_1$ and $I_2$ are similar to each other but dissimilar to $I_3$ and $I_4$. Second, in terms of property $p_1$, $I_1$ and $I_3$ are similar to each other but dissimilar to $I_2$ and $I_4$.

Keeping the first observation in mind, we take $(I_1, I_2, I_k)$ as a triplet of (query, positive, negative) images, where image $I_k$ is a potential answer image for $I_4$. We learn a feature weight vector $w^{c}_k$ to separate $I_1$ and $I_2$ from $I_k$. The superscript $c$ denotes that the query image $I_1$ and the positive image $I_2$ belong to the same category. If $I_k$ is a correct answer to the visual analogy question, $I_k$ needs to belong to the category $c_2$. This means that $w^{c}_{k}$ is learned to separate images belonging to category $c_1$ ($I_1$ and $I_2$) from images belonging to category $c_2$ ($I_k$). As $I_3$ belongs to category $c_2$, we can infer that if $I_k$ is correct, $w^{c}_k$ should be able to separate $I_1$ and $I_2$ from $I_3$. Thus we can use the extent to which $w^{c}_k$ can separate $I_1$ and $I_2$ from $I_3$ as a score to evaluate the possibility of $I_k$ being correct. Moreover, the score indicates the possibility of $I_k$ belonging to the same category $c_2$ with $I_3$.

Likewise, keeping the second observation in mind, we take $(I_1, I_3, I_k)$ as a triplet of (query, positive, negative) images. We learn a feature weight vector $w^{p}_k$ to separate $I_1$ and $I_3$ from $I_k$. The superscript $p$ denotes that the query image $I_1$ and the positive image $I_3$ have the same property. In the same way as described above, we can use the extent to which $w^{p}_k$ can separate $I_1$ and $I_3$ from $I_2$ as a score to evaluate the possibility of $I_k$ being correct. The score indicates the possibility of $I_k$ having the same property $p_2$ with $I_2$.

We have described two scores that can be used to find the correct answer image. To formalize this, we define the scoring function $S(w, q, p, n)$ to evaluate the extent to which a feature weight vector $w$ can satisfy a triplet of (query, positive, negative) images $(I_q, I_p, I_n)$, \ie, separate $I_q$ and $I_p$ from $I_n$:
\begin{align*}
S(w,q,p,n) = &\max\{0, ||Wx_q - Wx_p||^2_2 - \alpha_p\}+ \\
&\max\{\alpha_n - ||Wx_q - Wx_n||^2_2, 0 \}+\\
&\max\{\alpha_n - ||Wx_p - Wx_n||^2_2, 0 \}, \numberthis \label{score}
\end{align*}
where $W = \text{diag}(w_1, w_2, \ldots, w_d)$. The definition of $S(W,q,p,n)$ is exactly the same as the loss function \eqref{eq3} described above. Note that the lower $S(w,q,p,n)$ is, the better $w$ can satisfy the triplet $(I_q, I_p, I_n)$.

Given a visual analogy question $I_1:I_2::I_3:?$, for each potential answer image $I_k$, we propose to compute the score $S_k$ defined as follows
\begin{equation}
S_k = S(w^{c}_k, 1, 2, 3) + S(w^{p}_k, 1, 3, 2),
\label{scoreplus}
\end{equation}
where $w^{c}_k$ and $w^{p}_k$ denote the feature weight vector learned from the triplet $(I_1, I_2, I_k)$ and $(I_1, I_3, I_k)$ respectively. Once the scores are computed, we rank all the potential answer images accordingly.

\subsection{Unsupervised Attribute Discovery}
In this section, we describe our approach for unsupervised attribute discovery from the perspective of contextual visual similarity. Our goal is to discover cross-category attributes given only category labels of images.

A meaningful contextual visual similarity relationship, \ie, a triplet of (query, positive, negative) images, usually entails semantic attributes. For example, a triplet of black dog, black horse and white dog images entails the attribute `white'. In addition, if two triplets entail the same attribute, the learned feature weight vector of one triplet should be able to satisfy the other triplet. Based on the above observation, we propose our approach for attribute discovery.

Given a set of training images as well as category labels, we sample a large number of triplets of (query, positive, negative) images, where the query and negative images belong to the same category and the positive image belongs to a different category. It is expected that only a small fraction of the sampled triplets can make sense to human as we have no information to distinguish meaningless triplets from meaningful ones besides category labels.

Once we obtain the sampled triplets of images, we learn a feature weight vector $w_i$ for each triplet $i$. If triplet $i$ and $j$ entail the same attribute, the learned feature weight vectors $w_i$ and $w_j$ should be similar to each other, otherwise, they should be different. Therefore, we perform clustering on the learned feature weight vectors and hope that feature weight vectors grouped into the same cluster correspond to triplets that entail the same attribute. Note that we do not expect to model every triplet perfectly. Instead we expect that given enough data, some `easy' clusters emerge from the data and attributes are discovered.

Formally, our approach has three steps: (1) triplet sampling, (2) feature weight learning and (3) triplet clustering. During triplet sampling, for any two different categories $A$ and $B$, we select such pairs of images that the first image belongs to category $A$ and the second one belongs to category $B$ and the distance between their image features is smaller than a pre-set threshold $\theta_1$. The selected pairs of images are used as query and positive images. For a selected pair of images, we select $m$ furthest neighbor images to the query that also belong category A as negative images to form triplets of images. We denote the sampled triplets as $\{(q_i, p_i, n_i)\}$ where $q_i$, $p_i$ and $n_i$ are indices of query, positive and negative images respectively. We use the threshold $\theta_1$ to filter pairs of images as we hope the selected pairs of images have something (attributes) in common. We select furthest neighbor images to the query as we hope the negative images do not contain the same attribute with the query.

We then learn feature weight vector $w_i$ for each triplet $(q_i, p_i, n_i)$. After that, we employ Complete Linkage Clustering algorithm to cluster all the feature weight vectors $\{w_i\}$. Complete linkage clustering is an agglomerative hierarchical clustering algorithm and we stop clustering when the smallest distance between two clusters is larger than a threshold $\theta_2$. It is worthwhile to point out that we define a new distance metric between $w_i$ and $w_j$ instead of using Euclidean distance. The new distance metric $d(w_i,w_j)$ is defined to evaluate the extent to which $w_i$ can satisfy triplet $j$ and $w_j$ can satisfy triplet $i$. Here is the definition of $d(w_i,w_j)$:
\begin{equation}
d(w_i,w_j)=\max\{S(w_i, q_j, p_j, n_j), S(w_j, q_i, p_i, n_i)\},
\end{equation}
where the function $S(\cdot,\cdot,\cdot,\cdot)$ is defined in equation~\eqref{score}, $(q_i, p_i, n_i)$ represents triplet $i$ and $(q_j, p_j, n_j)$ represents triplet $j$. We use $\max$ operator here as we want to make sure that $w_i$ can satisfy triplet $j$ and $w_j$ can satisfy triplet $i$ at the same time.

As each feature weight vector corresponds to one triplet, we can obtain the clustering results of triplets based the clustering results of all the feature weight vectors $\{w_i\}$. We will show that the triplets entailing the same attribute tend to be grouped into the same cluster and meaningful clusters emerge from a large number of noisy sampled triplets.

\section{Experiments}

We evaluate our contextual visual similarity approach using triplet queries on three tasks: (1) attribute specific image search, (2) answering visual analogy questions and (3) unsupervised attribute discovery.

\subsection{Implementation Details}

In our experiments, the $L_2$ normalized activations from the pool5 or fc7 layer of VGGNet~\cite{simonyan2014very} are used as image features. They are referred as pool5 or fc7 in the following text.

We employ gradient descent algorithm to minimize the loss function and learn for the feature weight vector $w$. We set the learning rate to 0.1. $w$ is initialized to an all-one vector to treat each dimension equally at the beginning. The positive margin $\alpha_p$ and negative margin $\alpha_n$ are set to 0.5 and 2 respectively. $\lambda$ is set 1 to unless stated otherwise.

\subsection{Dataset}

A suitable dataset for us should consist of images of various object categories with distinguishable attributes. The attributes should also be shared across object categories so that attribute specific image search can be well evaluated or visual analogy questions can be constructed. We find the VAQA dataset used in ~\cite{sadeghi2015visalogy} most suitable for us. However the dataset was not available publicly so we collect our own version of the dataset following the methodolgy described in \cite{sadeghi2015visalogy}.

Inspired by~\cite{sadeghi2015visalogy}, we consider a list of object categories as well attributes and pair them to make a list of (category, attribute) labels. Uncommon combinations of categories and attributes like `red dog' are removed. In our experiments, we consider 8 categories, including bird, bus, car, cat, dog, horse, motorcycle and train, 8 attributes including black, blue, red, run, stand up, swim, white and yellow. In total, there are 38 common combinations of categories and attributes. We use each (category, attribute) label to query Google Image Search and download the returned images. After manually removing noisy images, we have 2197 images in total. On average, there exists about 60 images for each (category, attribute) combination.

\subsection{Attribute Specific Image Search Results}
We split the dataset into two subsets: query image set and database image set. For query image set, we randomly select 10 images from each (category, attribute) combination and this gives us 380 query images. All other 1817 images are used as database images. Since we have 8 attributes in total, on average there are roughly 225 correct database images for each attribute in the dataset. We show results where images are represented images by pool5 features and fc7 features respectively. The original pool5 and fc7 features are used as baselines.

To apply our approach, for a query image, we randomly sample $k$ positive images and $k$ negative images from the remaining images in query image set. In our experiments, we show results for $k = 1, 3$ and 5 respectively. Note that our approach is also applicable when there are different number of positive images and negative images. We set the number of positive images equaling to the number of negative images just for simplicity.

We employ Mean Precision and Mean Average Precision (MAP) as our evaluation metric. The database image is considered as correct if it contains the same attribute with the query. It is possible that a black dog is running but the image is only labeled as black in our dataset. A more comprehensive evaluation requires multi-label data, which we leave for future work.

\begin{figure}
\centering
\includegraphics[width=0.48\columnwidth]{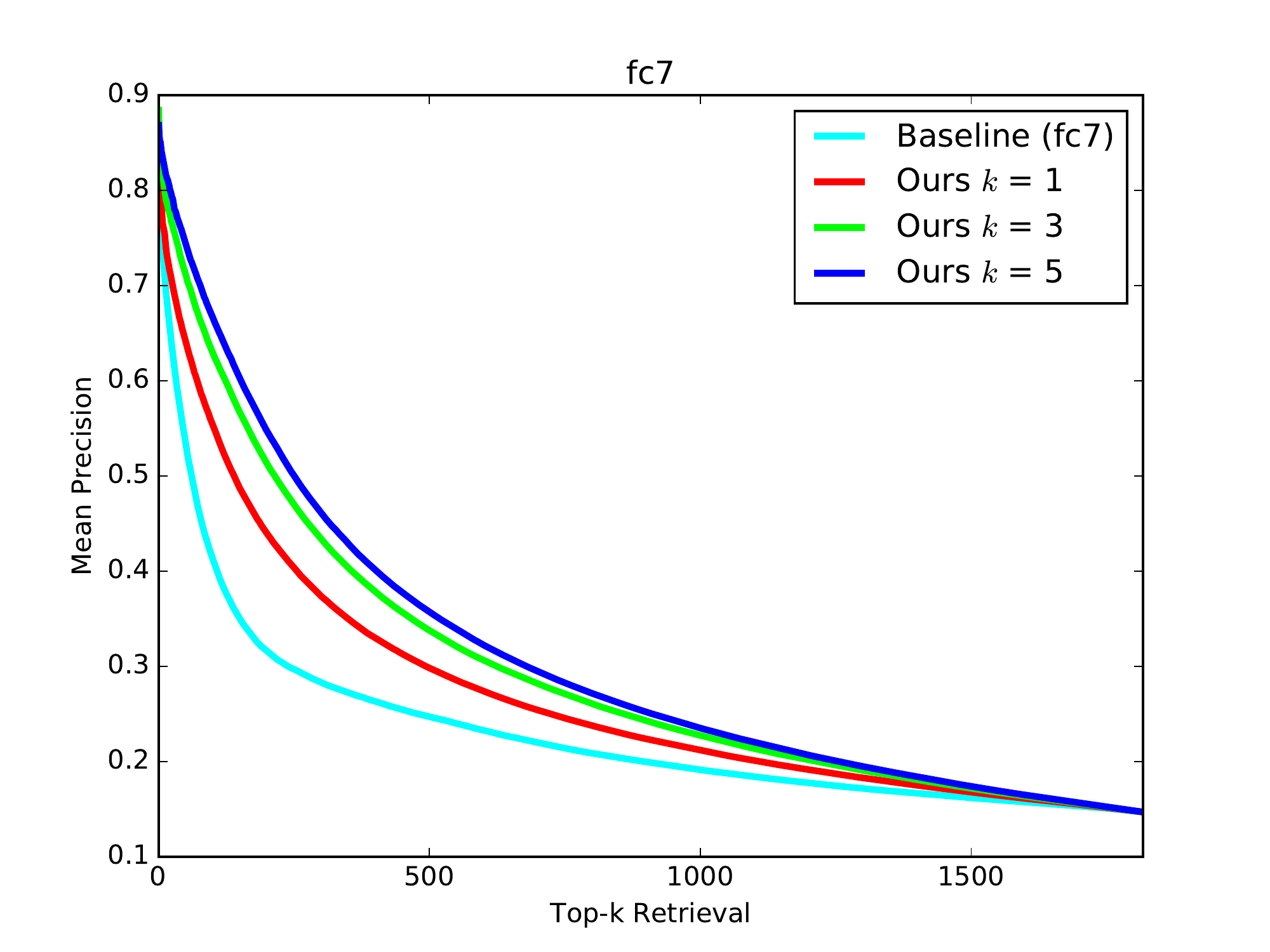}
\includegraphics[width=0.48\columnwidth]{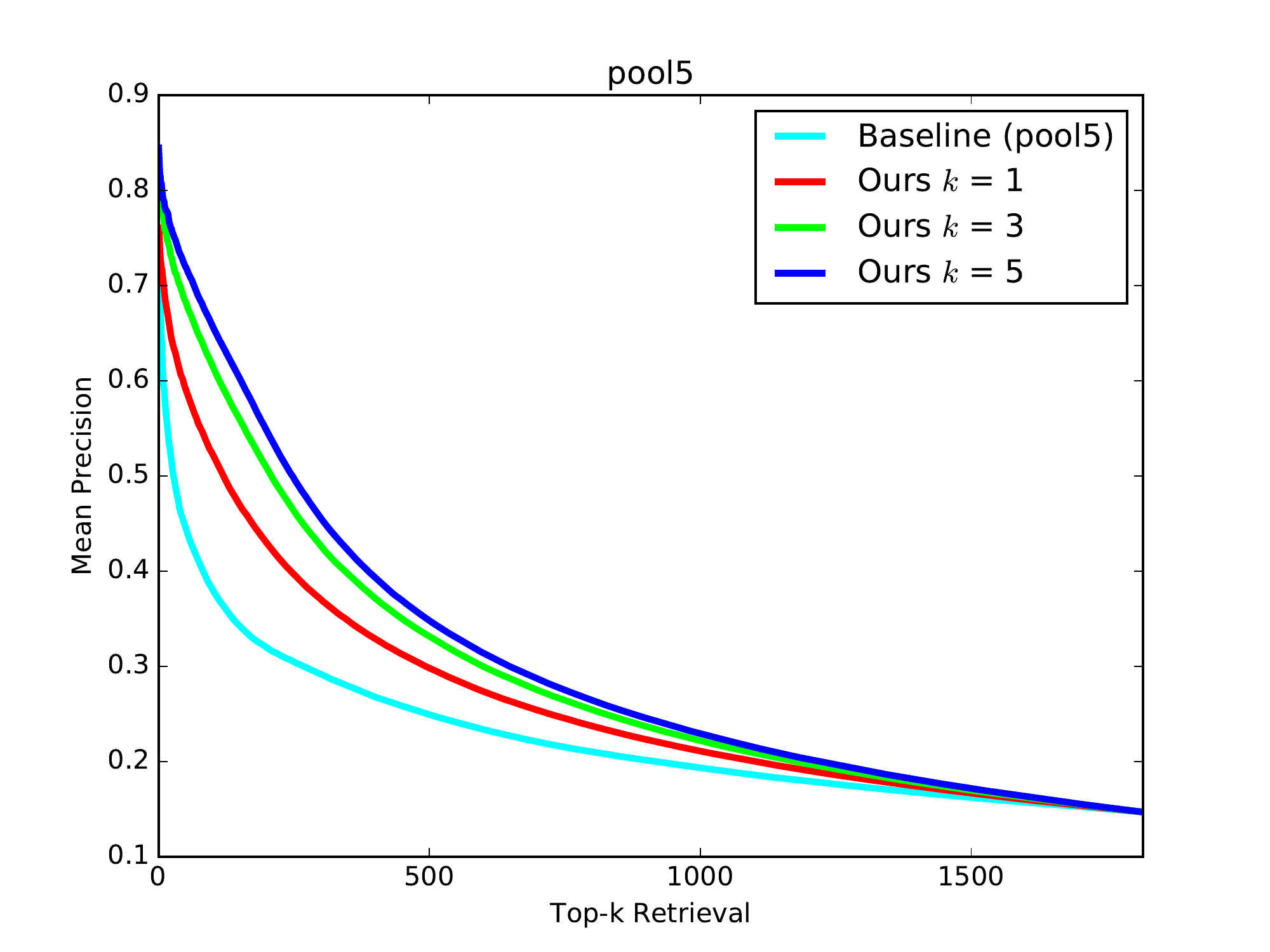}
\caption{Mean Precision of the baseline and our approach for attribute specific image search when using fc7 and pool5 features respectively.}
\label{fc7pool_pr}
\end{figure}

As shown in Table~\ref{MAP}, when using fc7 features, our approach significantly outperforms the baseline by about $11\%$, $18\%$ and $22\%$ when $k = 1, 3$ and $5$ respectively in terms of MAP. When using pool5 features, our approach also significantly outperforms the baseline. We can also see that as $k$ increases, \ie, the user provides more positive and negative images, the performance of our approach keeps improving. Mean precision results are shown Figure~\ref{fc7pool_pr}. In terms mean precision, our approach also outperforms the baseline.


\textbf{Ablation Study}: We study the influence of the parameter $\lambda$ on image search performance. In the following experiments, we use fc7 features to represent images and varying the value of $\lambda$. The results are reported in Table~\ref{lambda_MAP}. As shown in Table~\ref{lambda_MAP}, $\lambda = 0.1$ and $\lambda = 1$ yield the best results. When $\lambda = 0$, the performance is much worse than other settings. For example, when $k = 5$, the MAP for $\lambda = 0$ is lower than that for $\lambda = 1$ by about $20\%$, which proves the importance of the regularization term. We also notice that when $\lambda = 0$, more provided positive and negative images yield worse performance. Thus we can conclude that the regularization term can help avoid overfitting.



\subsection{Visual Analogy Results}

We randomly select 5 images from each (category, attribute) combination to form a set of potential answer images. In total we have 190 potential answer images. The remaining images are used to construct analogy questions.

We build analogy questions by instantiating the category and property in the general form~\eqref{quesform} of visual analogy questions. We choose the category from the 8 categories and the property from the 8 attributes in our dataset. Similar to~\cite{sadeghi2015visalogy}, we consider two types of analogy: color analogy and action analogy, where $p1$ and $p2$ are both color attributes or action attributes respectively. In total, we can build 428 types of instantiation of analogy questions with the shared properties (attributes) across categories. For each type of instantiation, we generate 10 questions so $4280$ analogy questions are generated altogether. For each question, there are 5 correct answers in the answer image set and other 185 images are served as distractor images. Note that when answering these questions, our approach only looks at the given images without being exposed to any category or property labels.

\textbf{Baseline:} We compare with the baseline approach used in~\cite{sadeghi2015visalogy}. We do not compare with the approach proposed in~\cite{sadeghi2015visalogy} as their approach requires training on a large number of labeled quadruples of images. However, our approach does not need any labels. The baseline approach uses the subtraction (difference) between image features to capture the mapping between images. Then they compare the mapping between $I_1$ and $I_2$ to the mapping between $I_3$ to a potential answer image to evaluate the possibility of the answer image being correct. Concretely, given a visual analogy question $I_1:I_2::I_3:?$, they rank all the potential answer images $\{I_k\}$ according to the following score:
\begin{equation}
R_{k} = \frac{T(I_1, I_2)\cdot T(I_3, I_k)}{||T(I_1, I_2)||\cdot||T(I_3, I_k)||},
\end{equation}
where $T(I_i, I_j)$ is defined as
\begin{equation}
T(I_i, I_j) = \frac{x_i-x_j}{||x_i - x_j||},
\end{equation}
where $x_i$ and $x_j$ denote the image feature of image $I_i$ and $I_j$ respectively. In our experiments, we use fc7 features to represent images for both the baseline and our approach.

\begin{figure}
\centering
\includegraphics[width=0.48\columnwidth]{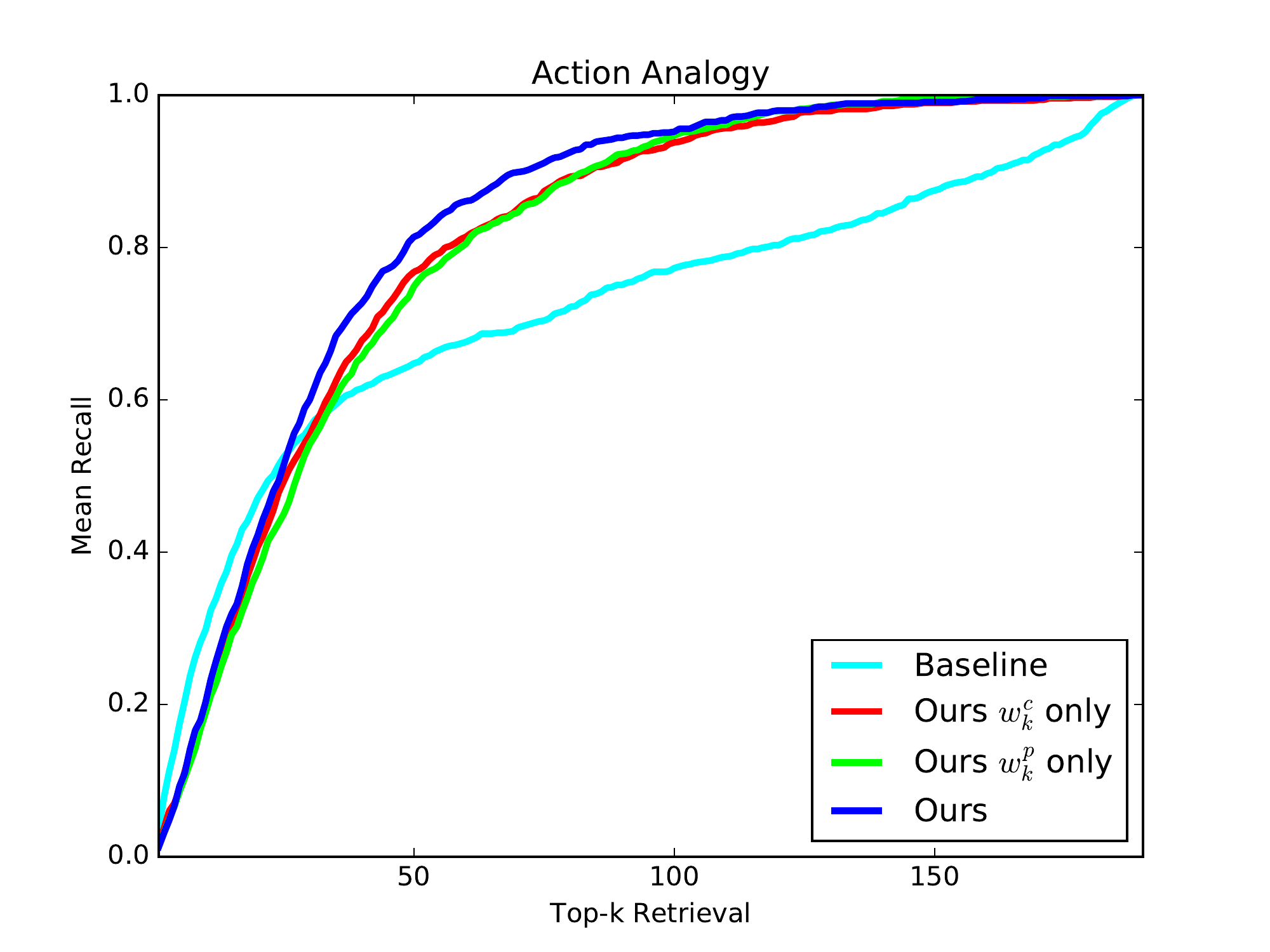}
\includegraphics[width=0.48\columnwidth]{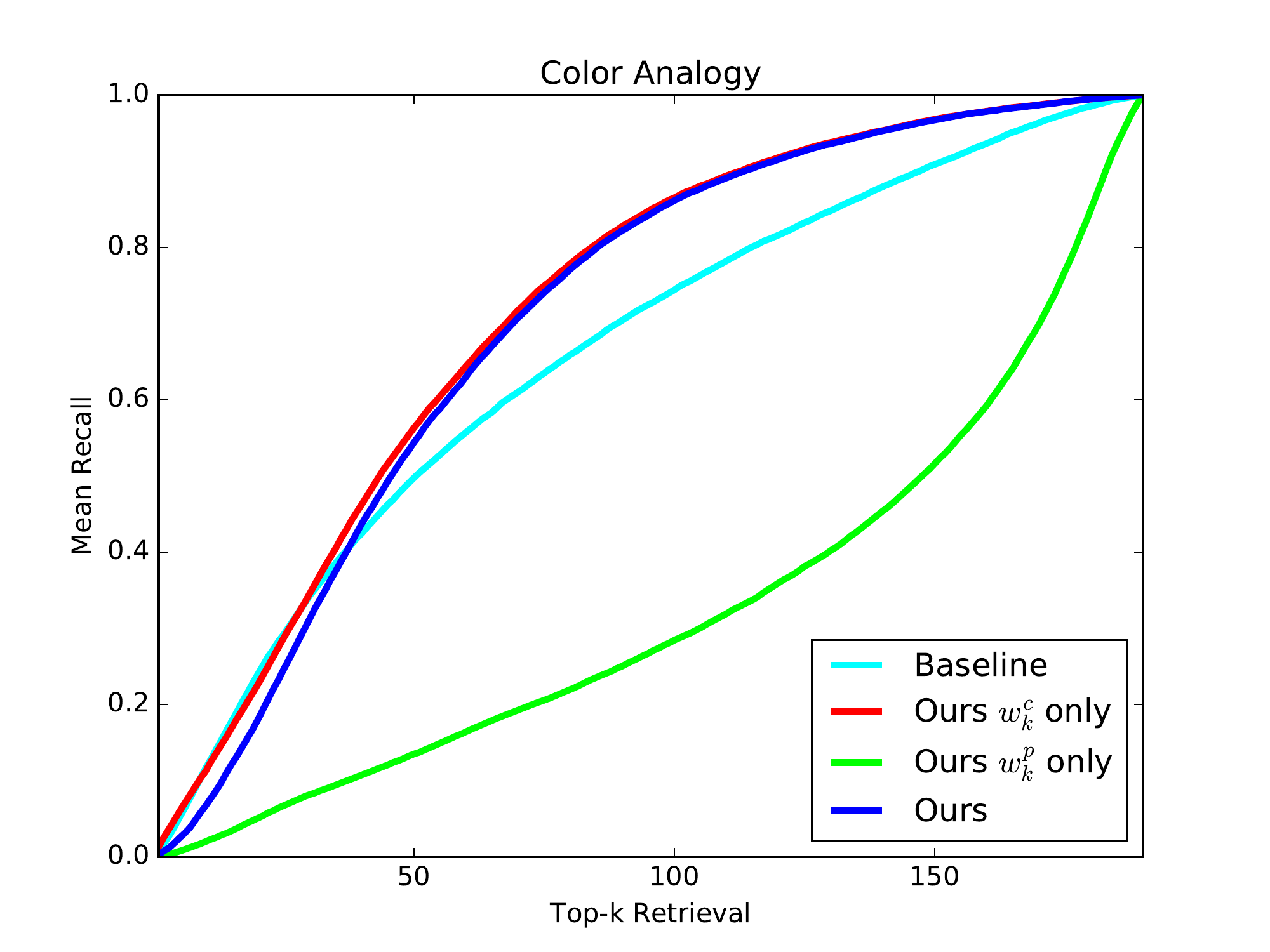}
\caption{Mean Recall of the baseline and our approach for action analogy and color analogy. `Ours $w^{c}_k$ only' means ranking answer images according to $S(w^{c}_k, 1, 2, 3)$ only. `Ours $w^{p}_k$ only' means ranking answer images according to $S(w^{p}_k, 1, 3, 2)$ only.}
\label{analogy_recall}
\end{figure}

\begin{figure*}
\centering
\includegraphics[width=1.72\columnwidth]{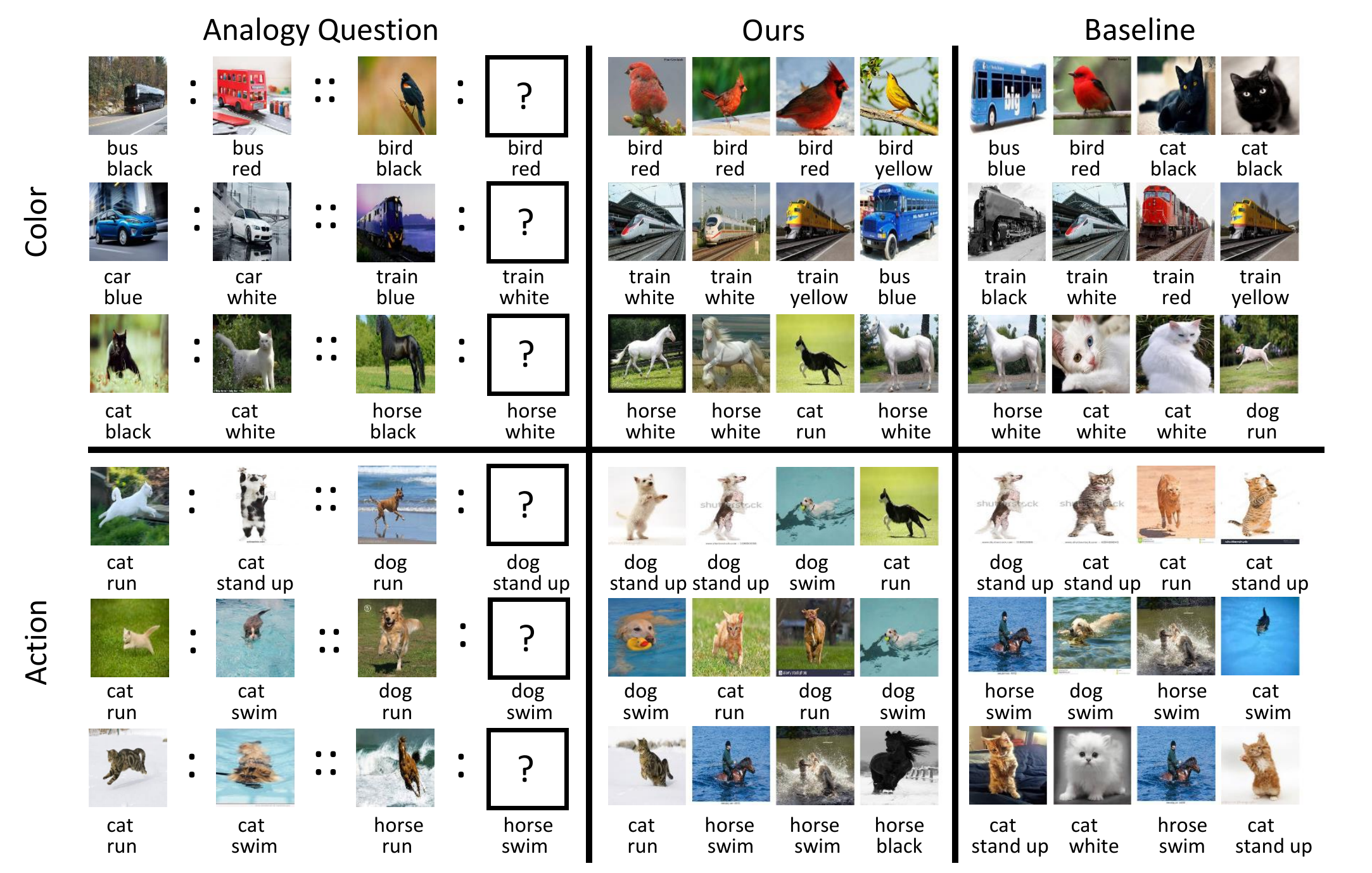}
\caption{Qualitative examples for answering visual analogy questions. The text under the image denotes the category and attribute label of the image. The text under the question mark image denotes the expected category and attribute label of the answer image. No labels are used when answering questions. We show labels here for illustration purpose only.}
\label{analogy}
\end{figure*}

Given a visual analogy question, both the baseline and our approach return a ranked list of answer images. Examples of visual analogy questions and top 4 returned images by the baseline and our approach are shown in Figure~\ref{analogy}. We use Mean Recall to evaluate the results quantitatively as~\cite{sadeghi2015visalogy} does. The performance is shown in Figure~\ref{analogy_recall}.

We can see that for both action analogy and color analogy questions, our approach can significantly outperform the baseline. For action analogy, ranking answer images according to $S(w^{c}_k, 1, 2, 3)$ only or $S(w^{p}_k, 1, 3, 2)$ only already yield better performance than the baseline. Combining them yields even better performance. However, for color analogy, ranking answer images according to $S(w^{p}_k, 1, 3, 2)$ yield bad performance. In the case of color analogy, the feature weight vector $w^{p}_k$ is learned to separate images with different colors. We conjecture that feature weight vectors learned to separate different colors are similar no matter they aim to separate black from white or black from yellow or so. This may lead to the failure of our approach when ranking according to $S(w^{p}_k, 1, 3, 2)$ only.

\begin{table}[tb]
\centering
\caption{Mean Average Precision (MAP) of the baseline and our approach for attribute specific image search when using fc7 and pool5 features respectively. The highest performance is shown in boldface.}
\label{MAP}
\begin{tabular}{|l|l|l|}
\hline
         & \multicolumn{1}{c|}{fc7} & \multicolumn{1}{c|}{pool5} \\ \hline
Baseline & 0.334                    & 0.320                      \\ \hline
$k$ = 1  & 0.440                    & 0.436                      \\ \hline
$k$ = 3  & 0.519                    & 0.522                      \\ \hline
$k$ = 5  & \textbf{0.557}                    & \textbf{0.565}                      \\ \hline
\end{tabular}
\end{table}

\begin{table}[tb]
\centering
\caption{Mean Average Precision (MAP) of our approach when varying the value of $\lambda$. We use fc7 features to represent images for the results shown in the table. The highest performance for each $\lambda$ is shown in boldface.}
\label{lambda_MAP}
\begin{tabular}{|c|l|l|l|l|l|}
\hline
\diagbox{$k$}{$\lambda$} & \multicolumn{1}{c|}{0} & \multicolumn{1}{c|}{0.01} & \multicolumn{1}{c|}{0.1} & \multicolumn{1}{c|}{1} & 10    \\ \hline
1 & \textbf{0.371} & 0.430 & 0.440 & 0.440 & 0.435 \\ \hline
3 & 0.365 & 0.498 & 0.520 & 0.519 & 0.491 \\ \hline
5 & 0.355 & \textbf{0.520} & \textbf{0.561} & \textbf{0.557} & \textbf{0.498} \\ \hline
\end{tabular}
\end{table}

\begin{figure*}
\centering
\includegraphics[width=1.72\columnwidth]{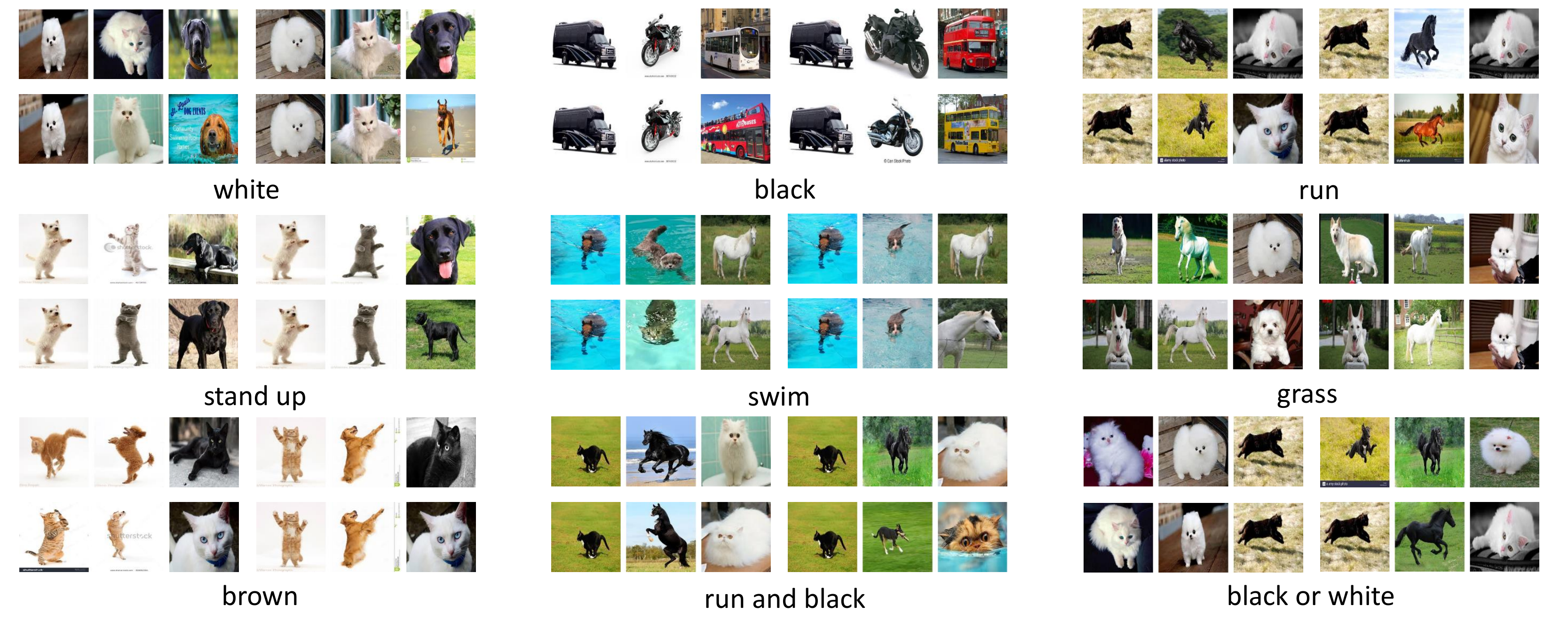}
\caption{Qualitative examples for attribute discovery. For each discovered attribute, we show 4 triplets of images from one cluster which is annotated as that attribute.}
\label{cluster}
\end{figure*}

\
\subsection{Unsupervised Attribute Discovery Results}

In the following experiments, we use fc7 features to represent images. The parameter $\theta_1$ is set to 1.1, $\theta_2$ is set to 1 and $m$ is set to 10. Under this parameter setting, 24420 triplets are sampled in total. After clustering, we obtain 116 cluster after removing small clusters whose size is smaller than 30.

To verify the quality of each cluster, we ask human annotators to assign an attribute to each cluster. They must answer the question, `What makes the first two images similar but but dissimilar to the third image, across all triplets belonging to that cluster?' If it is hard to name the attribute, the cluster will be annotated as a noisy cluster. In total, 72 clusters are considered as meaningful. Among them, 43 clusters are annotated as `white', 13 clusters are annotated as `black', 4 clusters are annotated as `run', 4 clusters are annotated as `stand up' and 1 cluster is annotated as `swim'. We also found that some clusters are annotated as attributes which are not present in the attribute vocabulary of our dataset. 2 clusters are annotated as `grass' and 1 cluster is annotated as 'brown'. We also find that there are clusters annotated as hybrid attributes. For example, 3 clusters are annotated as `run and black' and 1 cluster is annotated as `black or white'. Qualitative examples of the clustering results are shown in Figure~\ref{cluster}.

\section{Conclusion}

We have introduced the novel idea of modeling contextual visual similarity. Our approach learns a feature weight vector to encode contextual visual similarity. We have demonstrated that our approach can be used for three tasks: (1) attribute  specific  image  search,  (2)  answering visual analogy questions, and (3) unsupervised attribute discovery. We believe this work will help AI systems better understand different notions of similarity when the proper context is provided.

{\small
\bibliographystyle{ieee}
\bibliography{egbib}
}

\end{document}